\documentclass{article}
\usepackage{spconf,amsmath,graphicx,hyperref,subfig}
\title{\textbf{VOX-KRIKRI: Unifying Speech and Language through Continuous Fusion}}
\name{Dimitrios Damianos, Leon Voukoutis, Georgios Paraskevopoulos, Vassilis Katsouros }
\address{Institute for Speech and Language Processing, Athena Research Center, Greece}
\begin{document}
%
\maketitle
\begin{abstract}
We present a multimodal fusion framework that bridges pre-trained decoder-based large language models (LLM) and acoustic encoder-decoder architectures such as Whisper, with the aim of building speech-enabled LLMs. Instead of directly using audio embeddings, we explore an intermediate audio-conditioned text space as a more effective mechanism for alignment. Our method operates fully in continuous text representation spaces, fusing Whisper's hidden decoder states with those of an LLM through cross-modal attention, and supports both offline and streaming modes. We introduce \textit{VoxKrikri}, the first Greek speech LLM, and show through analysis that our approach effectively aligns representations across modalities. These results highlight continuous space fusion as a promising path for multilingual and low-resource speech LLMs, while achieving state-of-the-art results for Automatic Speech Recognition in Greek, providing an average $\sim20\%$ relative improvement across benchmarks.

\end{abstract}

\begin{keywords}
Speech LLMs, modality fusion, continuous latent space, causal masking, ASR
\end{keywords}
\section{Introduction}
\label{sec:intro}
Large language models (LLMs) have achieved remarkable success in natural language processing, inspiring the development of multimodal LLMs that combine pre-trained language models with modality-specific encoders such as those for audio and images~\cite{Qwen-VL,liu2023improved,wang2024cogvlm,ding2025kimi,radhakrishnan2023whispering}. This design departs from traditional end-to-end architectures~\cite{radford2021learning,elizalde2023clap} by leveraging the strengths of large pre-trained models.

Recent work in the speech domain can be broadly divided into approaches that use continuous speech representations and those that rely on discrete features. Continuous representations offer an intuitive and simple way to integrate speech signals into large language models (LLM). For example, in SLAM-ASR~\cite{ma2024embarrassingly}, continuous features are extracted from a speech decoder, downsampled, and then projected into the LLM’s embedding space. Similar strategies are employed by the Qwen-Audio series~\cite{chu2023qwen, chu2024qwen2} and SALMONN~\cite{tang2023salmonn}, where embeddings from the Whisper encoder~\cite{radford2023robust} and BEATs~\cite{chen2023beats} are combined via a Q-Former~\cite{li2023blip} and mapped into the LLM’s embedding space. Whispering-Llama~\cite{radhakrishnan2023whispering} takes a slightly different approach, inserting adapters and cross-attention layers into LLaMA~\cite{touvronllama} to fuse Whisper encoder embeddings with latent language representations. In contrast, discrete features do not require additional adapter modules for modality alignment and can be directly integrated into the LLM vocabulary. For instance, both SpeechGPT~\cite{zhang2023speechgpt} and AudioPaLM~\cite{rubenstein2021audiopalm} apply clustering techniques to discretize speech embeddings, while Kimi-Audio~\cite{ding2025kimi} uses a vector quantization layer to convert continuous representations into discrete token sequences.

Recent studies suggest that LLMs naturally operate in a continuous latent space. COCOMIX~\cite{tack2025llm} extracts continuous semantic concepts using a pre-trained Sparse Autoencoder (SAE)~\cite{shu2025survey} and identifies the most influential ones via attribution scores. Similarly, COCONUT~\cite{hao2024training} feeds the final continuous hidden LLM state directly as the input embedding for the next token, leveraging the full embedding space and enhancing reasoning capabilities. Additionally,~\cite{marrolanguage} shows that LLMs capture time- and space-continuous patterns and learn language in a continuous manner, in contrast with how humans understand language.

Motivated by these observations, we investigate whether leveraging an intermediate audio-conditioned text space can serve as a more effective mechanism for multimodal fusion compared to directly utilizing audio embeddings. To this end, we propose a novel multimodal fusion framework that operates entirely within the continuous representation space of both audio and language models. This approach is the first, to the best of our knowledge, to explore such continuous-space alignment for this purpose. Our contributions are: (1) we introduce a cross-modal adaptation framework that fuses Whisper’s decoder hidden states with a designated decoder layer of a large language model (LLM), enabling both offline and streaming fusion; (2) we develop VoxKrikri, the first Greek Speech LLM that achieves state-of-the-art ASR results across benchmarks; and (3) we conduct an analysis of the latent feature space, demonstrating that our framework successfully aligns the continuous representations of Whisper and the LLM. VoxKrikri will be made available under the Llama 3.1 Community License Agreement upon acceptance.




\begin{figure*}
    \centering
    \includegraphics[width=0.75\linewidth]{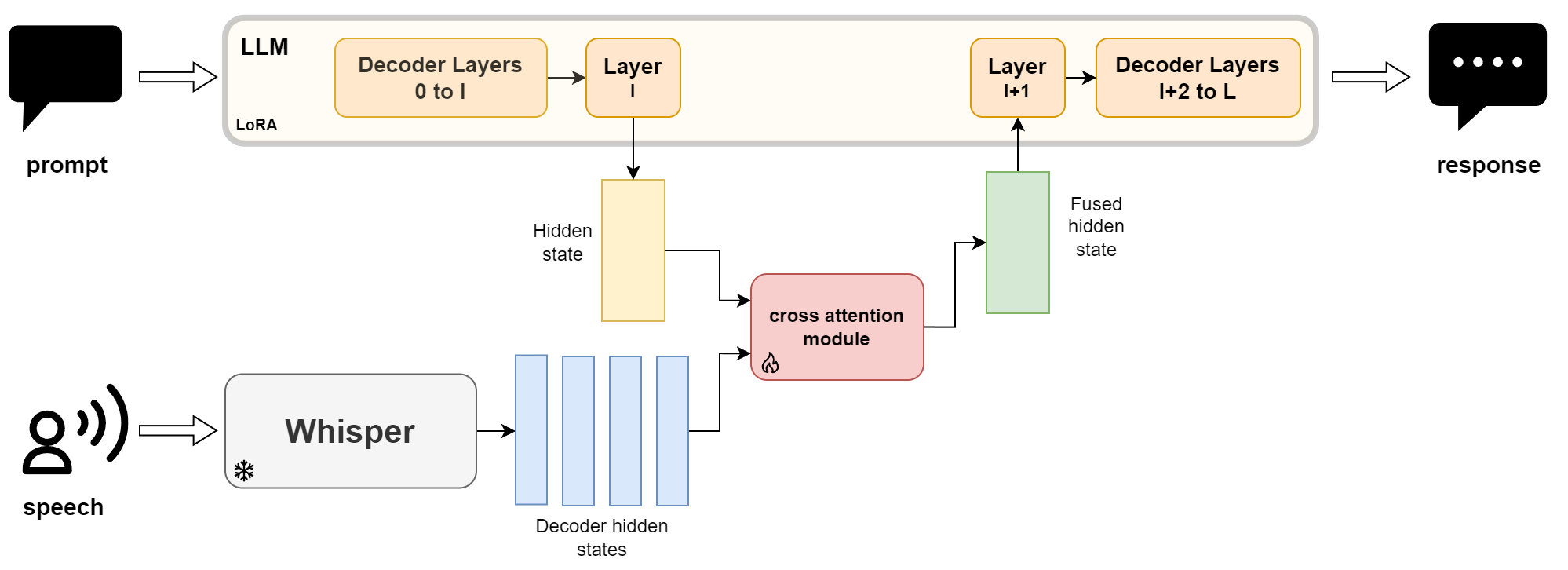}
    \caption{VoxKrikri: Whisper's decoder hidden states are fused with LLM's hidden states at a designated injection layer \textit{I}, using a cross-modality attention module. The attention module implements both full-sequence and causal fusion between the features, using a soft, proportional alignment method.}
    \label{fig:slm-framework}
\vspace{-1.5em}
\end{figure*}

\vspace{-0.5em}
\section{Proposed Method}
\label{sec:method}
We introduce a multimodal fusion approach for integrating audio and text features by using continuous hidden state spaces of the audio-conditioned decoder. An overview of the framework is shown in Fig.~\ref{fig:slm-framework}.

\subsection{Cross-modal Fusion}
We assume that the last hidden layer of Whisper’s decoder encodes audio-conditioned continuous textual representations, denoted as $A_s$. These are integrated into the LLM through a cross-attention module inserted at a decoder layer $I$, where text states $Y_t$ attend to audio-derived keys and values. Varying the insertion depth (early, intermediate, or late) allows us to study fusion at different semantic levels. Unlike approaches that project audio embeddings into the LLM input space~\cite{ma2024embarrassingly,chu2023qwen,tang2023salmonn}—which require the full audio sequence and thus only support offline use—our method enables both offline and streaming. Specifically, full-sequence fusion lets the LLM attend to the entire audio sequence, while causal fusion restricts attention to past frames for real-time applications.

\subsection{Full-sequence Fusion}
In full-sequence fusion, the entire audio sequence $A_{1:S}$ is fused with the text sequence $Y_{1:T}$ via the cross-attention module. The autoregressive capabilities of the LLM are preserved, since each prediction
\begin{equation}
    p(Y_{1:T}|A_{1:S}) = \prod_{t=1}^{T} p(Y_t|Y_{<t}, A_{1:S})
\end{equation}
depends on the preceding text tokens and the complete audio sequence. This approach is suitable for offline tasks where the full audio is available at inference time, such as audio question answering or speech summarization.

\subsection{Causal Fusion}
For real-time tasks such as streaming ASR or live captioning, temporal ordering must be preserved: each text token $Y_t$ may only attend to audio frames observed up to its decoding step:
\begin{equation}
    p(Y_t \mid Y_{<t}, A_{\le s_t}),
\end{equation}
where $s_t$ denotes the last audio frame aligned with token $t$. This design ensures that predictions are based only on past and current context, maintaining temporal consistency and the autoregressive nature of both Whisper and the LLM. During training and inference, a causal attention mask enforces this restriction.

\subsection{Causal Alignment Mask}
We implement a soft, proportional alignment strategy: each text token $t \in {0, \dots, T-1}$ attends only to audio tokens up to
\begin{equation}\label{align-position}
s_t = \Big\lfloor \frac{S}{T} \cdot t \Big\rfloor,
\end{equation}
where $T$ and $S$ are the numbers of text and audio tokens, respectively. This mapping enforces monotonicity: for any $t_1 < t_2$, $s_{t_1} \le s_{t_2}$, ensuring that later text tokens cannot access future audio frames. The proportional strategy also accommodates sequences of different lengths, providing roughly uniform coverage of audio frames across all text tokens. The causal attention mask $M \in {R}^{T \times S}$ is then defined as

\begin{equation}\label{causal-mask} 
M_{t,s} = \begin{cases} 0, & s \le s_t \\ -\infty, & s > s_t \end{cases}, \quad 0 \le t \le T-1, \quad 0 \le s \le S-1, 
\end{equation}
\noindent
and is applied directly to the attention logits before softmax. This formulation allows flexible fusion strategies: setting $s_t = S-1$ for all $t$ recovers full-sequence fusion, while Eq.~\eqref{align-position} enforces causal fusion. Finally, the fused hidden state of the LLM passed to layer $I+1$ is computed as:
\begin{equation}\label{cross-modal-fusion}
h_t = h_t + \text{softmax}\Big( \frac{(Y_tQ)(A_sK)^T}{\sqrt{d}} + M_{t,s} \Big) (A_s V).
\end{equation}
where $Q,K,V$ the query, key and value matrices of the cross attention module, respectively. During training these parameters are fully trainable.

\vspace{-0.5em}
\section{Experimental Setup}
\subsection{Speech Model}
We adopt Whisper-large-v3 as the acoustic backbone. This encoder–decoder model achieves state-of-the-art results on Greek ASR benchmarks and serves as a robust foundation. During training, Whisper parameters remain frozen.

\subsection{LLM and LoRA Adaptation}
For the language backbone, we employ Llama-KriKri-8B~\cite{roussis2025krikri}, a Greek-adapted version of Llama 3.1-8B, which substantially outperforms the base Llama 3.1-8B on Greek benchmarks.\footnote{\url{https://huggingface.co/collections/ilsp/ilsp-greek-evaluation-suite-6827304d5bf8b70d0346b02c}} We apply LoRA~\cite{hu2022lora} adapters for parameter-efficient adaptation during training. Regarding the injection layer $I$ of the LLM, whose hidden states are fused with audio features, we experiment with the 1st, 9th, 15th, 21st and 30th layers out of the 32 decoder layers of Llama 3.1-8B. Each configuration is denoted by its corresponding model name: \textit{VoxKrikri-1}, \textit{VoxKrikri-9}, \textit{VoxKrikri-15}, \textit{VoxKrikri-21}, and \textit{VoxKrikri-30}.\\

\vspace{-0.5em}
\begin{table}[h!]
    \centering
    \begin{tabular}{l|r|r}
    \textbf{Dataset} & \textbf{\#Hours}  & \textbf{\#Samples} \\
    \hline 
    \hline
    GPC-2400 & 2447 & 430K \\
    GPC-50 & 800 &  488K  \\
    Fleurs & 13  & 3,2K \\
    CV & 12  & 10,8K \\
    LG & 72  & 23,5K \\
    HParl & 120  & 76K \\
    \hline
    Total & $\sim$3300 & $\sim$1M \\
    \end{tabular}
    \caption{Training data analysis}
    \label{tab:training-data}

\vspace{-1.5em}
\end{table}

\subsection{Datasets}
For training, we employed the following datasets:\

\noindent\textbf{Logotypografia (LG)~\cite{digalakis2003large}} is one of the earliest Greek corpora, comprising approximately 72 hours of speech for training and 9 hours for testing.

\noindent\textbf{Common Voice (CV)~\cite{ardila2020common}} is a multilingual, crowd-sourced dataset developed by Mozilla. In our experiments, we use version 9.0 of the Greek subset, which contains 12 hours of training data, and 2 hours of test data.

\noindent\textbf{HParl (HP)~\cite{paraskevopoulos2023sample}} consists of parliamentary recordings from the Hellenic Parliament, which consists of 120 hours of training and 11 hours of testing data.

\noindent\textbf{Fleurs~\cite{conneau2023fleurs}} is a multilingual speech corpus. The Greek portion comprises 13 hours of speech for training and 2 hours for testing.

\noindent\textbf{Greek Podcast Corpus (GPC)~\cite{paraskevopoulos2024greek}} is a weakly supervised dataset of Greek podcasts, covering 16 diverse  categories (e.g., \textit{True Crime}, \textit{Comedy}). We use the GPC-50 subset for training, containing 50 hours per category (800 hours total), and a test set of 1 hour per category (16 hours total). In addition, we collected 2,447 hours of transcribed Greek podcast audio (annotated as \textit{GPC-2400}), totaling 434,530 samples. Table~\ref{tab:gpc-1400} details its category distribution, while Table~\ref{tab:training-data} overviews the full training data.

\begin{table}[h!]
    \centering
    \begin{tabular}{l|r|r}
    \textbf{Category} & \textbf{\#Hours}  & \textbf{\#Samples} \\
    \hline 
    \hline
    Arts & 361 &  62K  \\
    Business & 192  & 33K \\
    Comedy & 263  & 44K \\
    Education & 417  & 70K \\
    Fiction & 290  & 49K \\
    Health & 312  & 52K \\
    History & 170  & 48K \\
    Kids & 144  & 28K \\
    Leisure & 130  & 20K \\
    Music & 168  & 28K \\
    \hline
    Total & $\sim$2400 & $\sim$430K \\
    \end{tabular}
    \caption{GPC-2400 categories breakdown}
    \label{tab:gpc-1400}
\vspace{-1.5em}
\end{table}


\begin{table*}[t!]
    \centering
    \begin{tabular}{l||cc|cc|cc|cc|cc||c}
    \textbf{Dataset} 
      & \multicolumn{2}{c|}{\textbf{VoxKrikri-1}} 
      & \multicolumn{2}{c|}{\textbf{VoxKrikri-9}} 
      & \multicolumn{2}{c|}{\textbf{VoxKrikri-15}} 
      & \multicolumn{2}{c|}{\textbf{VoxKrikri-21}} 
      & \multicolumn{2}{c||}{\textbf{VoxKrikri-30}} 
      & \textbf{Whisper-large-v3} \\
   
    & causal & full
    & causal & full
    & causal & full
    & causal & full
    & causal & full
    & \\ 
    \hline \hline
    GPC & 14.88 & 14.81 & 13.06 & 14.26 & 12.65 & 12.75 & \underline{12.27} & \textbf{12.08} & 12.64 & 12.6 & 17.27 \\
    \hline
    Fleurs & 10.85 & 10.5 & 11.13 & 10.21 & 10.2 & 10.3 & 9.52 & \textbf{9.33} & \underline{9.48} & 10.2 & 11.02 \\
    \hline
    CV & 13.8 & 13.6 & 12.42 & 12.2 & 12.3 & \textbf{11.97}  & \underline{12.01} & 12.04 & 12.43 & 12.1 & 13.98 \\
    \hline
    LG &10.55& 10.41 & 10.03 & 9.89 & 9.78 & 9.54 & \underline{9.47} & \textbf{9.4} & 9.66 & 9.52 & 10.86 \\
    \hline
    HParl & 14.7  & 15.01 & 13.9 & 13.8 & 13.55 & 13.32 & \textbf{12.9} & \underline{13.01} & 13.41 & 13.3 & 16.99 \\
    \hline
    \end{tabular}
    \caption{ASR results on benchmark test sets (WER) using both causal and  full-sequence fusion}
    \label{tab:wer-results}
\vspace{-2.5em}
\end{table*}

\begin{figure*}[t!]
    \centering
    \subfloat[]{\includegraphics[width=0.5\textwidth]{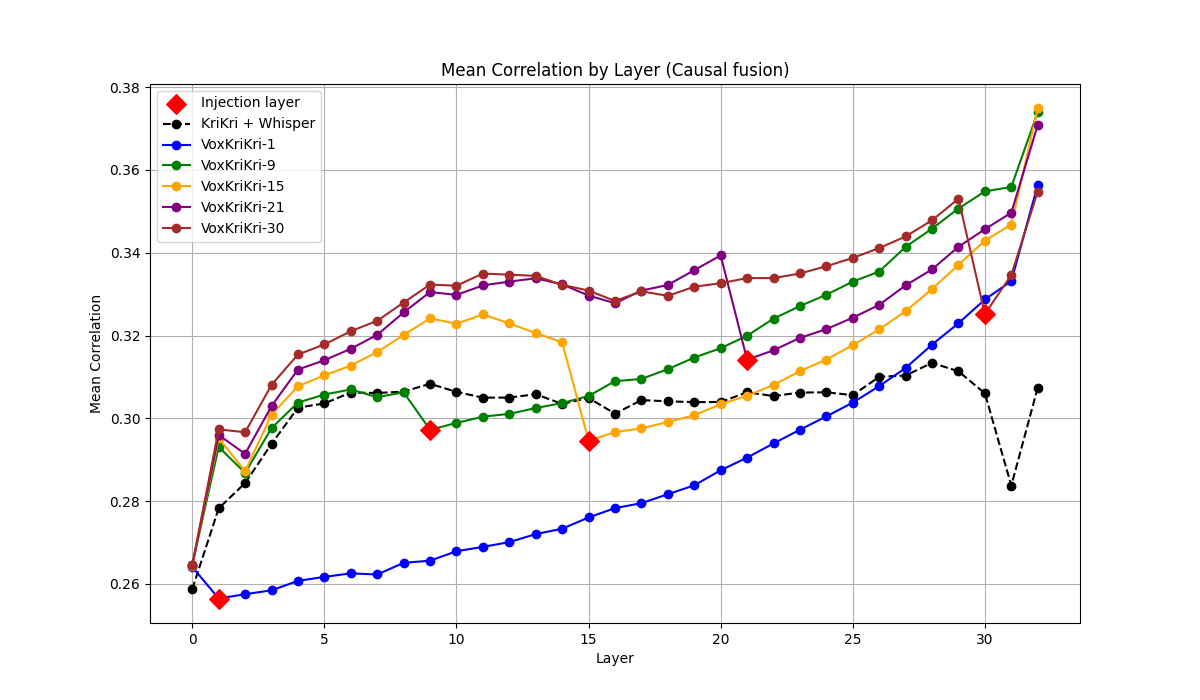}\label{fig:a}}
    \subfloat[]{\includegraphics[width=0.5\textwidth]{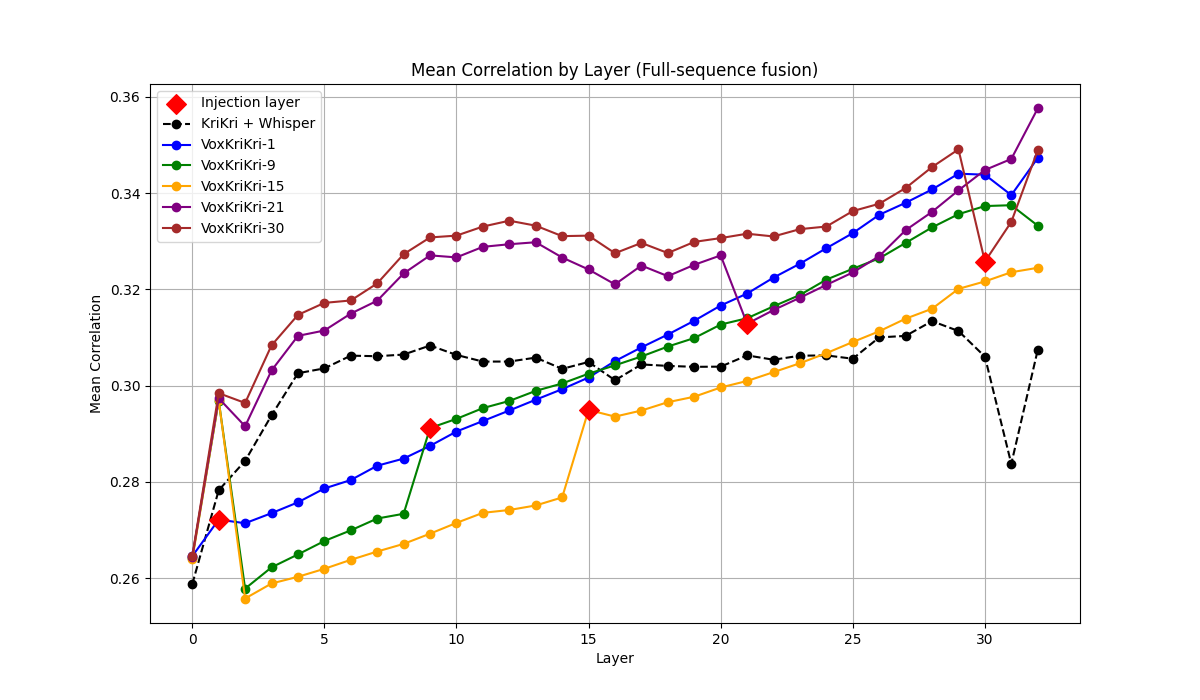}\label{fig:b}}
    \caption{Mean CCA correlation per LLM layer between the hidden states and the Whisper decoder features for the base models and VoxKrikri-1,9,15,21,30. (a) Using causal fusion, (b) Using full-sequence fusion. Overall, intermediate-to-late fusion increases cross-modal correlation. }
    \label{fig:cca-result}
\vspace{-1em}
\end{figure*}

\vspace{-0.5em}
\subsection{Experimental setting}
For fine-tuning with the LoRA adapter, we set the hyperparameters to $r=8$, $\alpha=16$, and $dropout=0.1$. This configuration resulted in approximately $744$ million trainable parameters out of a total of $9.8$ billion (around $7.53\%$ of the full model). All experiments were conducted on NVIDIA A100 GPUs with 64GB memory, provided by the LEONARDO supercomputer~\cite{turisini2024leonardo}.

\vspace{-0.5em}
\section{Results}
We evaluate the \textit{VoxKrikri-1,9,15,21,30} models on Automatic Speech Recognition (ASR) using Word Error Rate (WER) and compare them against the state-of-the-art Whisper-large-v3 baseline across five test sets (Table~\ref{tab:wer-results}). Both full-sequence and causal fusion strategies are considered. Across all benchmarks, every \textit{VoxKrikri} variant consistently outperforms Whisper, with improvements ranging from ~0.6 to over 5 WER points depending on the dataset. Full-sequence fusion generally yields slightly lower WER than causal fusion, as seen for example on GPC-test (12.08 vs. 12.27) and Fleurs-test (9.33 vs. 9.52). Moreover, intermediate-to-late fusion layers (15, 21, and 30) tend to outperform early fusion (1, 9), with \textit{VoxKrikri-21} achieving the best overall results on three datasets (GPC, Fleurs, and LG) and \textit{VoxKrikri-15} excelling on CV-test. Notably, \textit{VoxKrikri-21 (full)} reduces WER by over 5 points on GPC-test compared to Whisper (12.08 vs. 17.27). These results highlight the robustness of our continuous latent fusion approach and demonstrate that carefully chosen fusion depths enable substantial performance gains.


\vspace{-0.5em}
\section{Cross-modal Alignment}
To study cross-modal alignment, we measure correlations between the LLM’s layer-wise representations (encoding transcriptions) and Whisper’s decoder features (processing speech) using rCCA~\cite{tuzhilina2023canonical} on 1,000 samples from the GPC test set, shown in Fig.~\ref{fig:cca-result}. To address dimensionality mismatch, we subsample 20,000 features and run rCCA with 1,000 components per view and regularization $\lambda=10^{-4}$. Results show that intermediate-to-late fusion consistently increases correlation, strengthening representational alignment between modalities. Additionally, the sudden increases and decreases indicate that early-to-intermediate fusion benefits from access to the entire sequence, while late fusion is less sensitive to it. Overall, intermediate-to-late fusion achieves the strongest alignment, in line with the WER improvements observed.


\vspace{-0.75em}
\section{Conclusion \& Future Work}
In this work, we proposed a novel framework for narrowing the modality gap between pre-trained language and acoustic models by operating directly in their textual, continuous embedding spaces. Rather than relying on raw audio embeddings, our approach leverages an intermediate audio-conditioned text space, providing a linguistically grounded representation for more effective multimodal fusion. Building on this framework, we introduced \textit{VoxKrikri}, the first Greek Speech-LLM, which achieves state-of-the-art performance on ASR tasks. Our analysis further showed that the proposed methodology effectively improves the alignment between speech and language representations. We believe this framework offers a strong foundation for future research on multimodal alignment in continuous latent spaces. Moreover, our introduction of causal fusion via causal cross-modal masking paves the way for streaming and real-time applications of SpeechLLMs.

Looking ahead, we plan to extend our work in several directions. First, we aim to incorporate general-purpose audio capabilities, for instance by integrating a BEATs encoder, and to evaluate our framework on more challenging tasks such as speech and audio question answering, or real-time translation. In addition, we are interested in exploring multi-layer injection strategies and experimenting with combinations of early and late fusion.

\vspace{-0.5em}
\section{Acknowledgments}
We acknowledge the EuroHPC Joint Undertaking for awarding this project access to the EuroHPC supercomputer LEONARDO, hosted by CINECA (Italy) and the LEONARDO consortium through an EuroHPC Development Access call.

\bibliographystyle{IEEEbib}
\bibliography{refs}

\end{document}